\documentclass[10pt,twocolumn,letterpaper]{article}

\usepackage[pagenumbers]{cvpr} 

\newcommand{\TODO}[1]{\textbf{\color{red}[TODO: #1]}}
\usepackage{pifont} 
\usepackage{multirow}
\usepackage{tabularx}
\usepackage{ragged2e}
\usepackage{setspace}  
\usepackage[export]{adjustbox}
\usepackage[table]{xcolor} 
\usepackage{makecell}

\usepackage{graphbox} 
\usepackage{array}    
\usepackage[accsupp]{axessibility}  

\renewcommand{\TODO}[1]{}

\usepackage{microtype}

\renewcommand{\paragraph}[1]{\vspace{.5em}\noindent\textbf{#1.}}

\usepackage{xspace}

\usepackage{soul}
\setuldepth{foobar}

\definecolor{cvprblue}{rgb}{0.21,0.49,0.74}
\usepackage[pagebackref,breaklinks,colorlinks,allcolors=cvprblue]{hyperref}

\def\paperID{61} 
\def\confName{CVPR}
\def\confYear{2026}

\title{R3PM-Net: Real-time, Robust, Real-world Point Matching Network}

\author{
Yasaman Kashefbahrami$^1$ \quad Erkut Akdag$^1$ \quad Panagiotis Meletis$^2$ \quad Evgeniya Balmashnova$^2$ \\
Dip Goswami$^1$ \quad Egor Bondarau$^1$ \\
$^1$\textit{Department of Electrical Engineering, Eindhoven University of Technology} \quad $^2$\textit{Sioux Technologies}\\
}

\begin{document}
\maketitle
\begin{abstract}

\textit{Accurate Point Cloud Registration (PCR) is an important task in 3D data processing, involving the estimation of a rigid transformation between two point clouds. While deep-learning methods have addressed key limitations of traditional non-learning approaches, such as sensitivity to noise, outliers, occlusion, and initialization, they are developed and evaluated on clean, dense, synthetic datasets (limiting their generalizability to real-world industrial scenarios). This paper introduces R3PM-Net, a lightweight, global-aware, object-level point matching network designed to bridge this gap by prioritizing both generalizability and real-time efficiency. To support this transition, two datasets, Sioux-Cranfield and Sioux-Scans, are proposed. They provide an evaluation ground for registering imperfect photogrammetric and event-camera scans to digital CAD models, and have been made publicly available. Extensive experiments demonstrate that R3PM-Net achieves competitive accuracy with unmatched speed. On ModelNet40, it reaches a perfect fitness score of $1$ and inlier RMSE of $0.029$ cm in only $0.007$s, approximately $7\times$ faster than the state-of-the-art method RegTR~\cite{Yew2022REGTR:Transformers}. This performance carries over to the Sioux-Cranfield dataset, maintaining a fitness of $1$ and inlier RMSE of $0.030$ cm with similarly low latency. Furthermore, on the highly challenging Sioux-Scans dataset, R3PM-Net successfully resolves edge cases in under 50 ms. These results confirm that R3PM-Net offers a robust, high-speed solution for critical industrial applications, where precision and real-time performance are indispensable. The code and datasets are available at \url{https://github.com/YasiiKB/R3PM-Net}.}


\end{abstract}    
\section{Introduction}
\label{sec:intro}


\begin{figure}[htbp]
\centering
\includegraphics[width=0.4\textwidth]{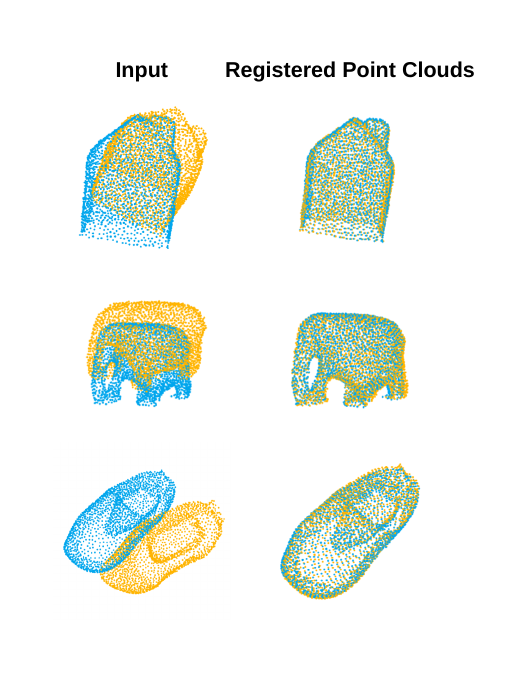} 
\caption{Qualitative results of R3PM-Net. As a real-time, feature-based deep learning method for point cloud registration, R3PM-Net improves generalizability and inference speed by expanding the network’s field-of-view.}
\label{fig:intro}
\end{figure}

A point cloud is a set of three-dimensional (3D) data points representing the external surface of objects or environments~\cite{Li2021AChallenges}. Rapid advances in sensor technology have made the acquisition of these data increasingly accessible for various applications in biomedical imaging, robotics, and automotive. In these domains, Point Cloud Registration (PCR) is a fundamental yet challenging process that aims to estimate a rigid transformation (i.e. rotation and translation) that aligns two point clouds. Accurate registration is the foundation for downstream tasks, such as 3D reconstruction, simultaneous localization and mapping (SLAM), and automated quality inspections~\cite{Zhang2024ALearning}.

Traditional non-learning methods for PCR, such as Iterative Closest Point (ICP)~\cite{Besl1992AShapes} and Random Sample Consensus (RANSAC)~\cite{Fischler1981RandomConsensus} struggle with high noise, outliers, incomplete scans, and are highly sensitive to initial pose estimation. To address these limitations, learning-based approaches have been proposed, demonstrating improved accuracy, robustness, and efficiency~\cite{Yew2020RPM-Net:Features, aoki2019pointnetlk, Wang2019DeepRegistration}. 

However, despite the abundance of deep-learning methods, most approaches are limited to synthetic datasets for both training and evaluation. Such datasets often fail to capture the complexity of real-world data, including noise, occlusions, sparsity, and incomplete coverage~\cite{Denayer2024ComparisonObjects, Fontana2020AAlgorithms}. As a result, their generalization to real industrial scenarios remains restricted~\cite{Zhang2020DeepOverview}. To address this gap in existing datasets, this work contributes two collections; the \textit{Sioux-Cranfield dataset}, consisting of both perfect and challenging 3D computer-aided design (CAD) models and the \textit{Sioux-Scans} dataset, which addresses the real-world challenge for registration of noisy, occluded, and sparse event-camera scans to digital CAD models. 

Moreover, current state-of-the-art approaches typically rely on hybrid feature representations~\cite{Yew2020RPM-Net:Features, Deng2018PPFNet:Matching, Slimani2024LoGDesc:Registration} that integrate local geometric primitives with global context through high-dimensional embeddings. Alternatively, they employ complex backbones~\cite{Huang2020PREDATOR:Overlap, Qin2022GeometricRegistration, Yew2022REGTR:Transformers} to combine global positional awareness and local descriptors. While being effective on dense synthetic datasets, these features often lack robustness in sparse real-world scenarios. For instance, in event-camera scans, under-populated neighborhoods provide insufficient data for reliable semantic information. Furthermore, the high computational overhead of integrating features or employing heavy backbones introduces significant latency, limiting applicability in time-critical industrial settings, such as real-time in-line quality control. 

To overcome these challenges, R3PM-Net, a \textbf{R}eal-time, \textbf{R}obust, and \textbf{R}eal-world data-focused \textbf{P}oint \textbf{M}atching Network, proposes a lightweight, global-aware feature extraction module. Instead of restricting the network to local neighborhoods or relying on increasingly complex backbones, R3PM-Net adopts a simplified design that expands the effective receptive field to capture a broader geometric context. Therefore, R3PM-Net offers two key advantages:

\begin{itemize}
    \item \textit{Global Context Awareness:} A broader 3D receptive field aggregates global information, producing robust descriptors even with occluded or sparse local data. 
    \item \textit{Real-Time Efficiency:} Eliminating expensive feature engineering and heavy backbones reduces inference latency to below 50 milliseconds (ms), enabling deployment in time-critical industrial applications.
\end{itemize}

Extensive experiments on both the publicly available and the introduced datasets indicate that R3PM-Net achieves state-of-the-art results by a minimalist architecture with significant efficiency ($50$ ms). R3PM-Net matches the performance of complex, feature-extraction-based models on synthetic datasets with a fraction of their complexity and runtime, while improving accuracy on real-world data. To summarize, the main contribution of this work is as follows.

\begin{itemize}

\item R3PM-Net, a robust point matching network for real-world object-level applications, that employs a lightweight, global feature extraction architecture to handle sparse, noisy, incomplete inputs.

\item The Sioux-Cranfield dataset, to bridge the gap between ``pristine" CAD models and ``noisy" photogrammetric reconstructions and enable evaluation across varying data quality levels.

\item The Sioux-Scans dataset, for registering sparse, and occluded event-camera scans to CAD targets. This dataset highlights the performance of R3PM-Net in practical industrial scenarios, where sensor noise and occlusions are present and precise ground-truth poses are unavailable.

\item Comprehensive evaluations demonstrate that R3PM-Net matches the state-of-the-art performance on synthetic benchmarks with significantly faster inference. On the challenging realistic data, R3PM-Net maintains real-time latency while consistently outperforming more complex models.
\end{itemize} 

\section{Related Work}
\label{sec:related}

Point Cloud Registration (PCR) can be broadly categorized into non-learning and learning-based approaches~\cite{Zhang2024ALearning, Zhang2020DeepOverview, Chen2024ALearning, Huang2021ARegistration}. While traditional methods rely on geometric optimization, recent deep learning approaches learn feature representations and enable end-to-end optimization through differentiable alignment.

\subsection{Non-learning Methods}
Traditional approaches primarily focus on iterative optimization. The Iterative Closest Point (ICP) algorithm~\cite{Besl1992AShapes} and its variants (e.g., Point-to-Plane~\cite{Yang1992ObjectImages}, Generalized-ICP~\cite{Segal2010Generalized-ICP}) minimize spatial distances between corresponding points. However, they are sensitive to initialization and prone to convergence to local minima. To mitigate outliers, RANSAC~\cite{Fischler1981RandomConsensus} employs a hypothesize-and-verify scheme, but its iterative nature limits scalability and real-time performance. 

\subsection{Learning-based Methods}
Deep learning methods replace hand-crafted descriptors of traditional approaches with learned features that capture local geometric structure. 
To this end, existing \textit{projection-based} approaches~\cite{Su2015Multi-viewRecognition, Zhu2014DeepRetrieval, Wu2023PointStronger} map 3D points to 2D images for 2D CNN processing, but often lose geometric details. \textit{Voxel-based} methods~\cite{Zeng20173DMatch:Reconstructions, Huang2020PREDATOR:Overlap} voxelize point clouds into volumetric grids for 3D convolutions at the cost of high memory consumption and quantization artifacts. \textit{Point-based} architectures~\cite{Qi2016PointNet:Segmentation, Deng2018PPFNet:Matching, Wang2018DynamicClouds} operate directly on raw point sets and preserve geometric detail. Recent state-of-the-art models, such as KPConv~\cite{Thomas2019KPConv:Clouds} in GeoTransformer~\cite{Qin2022GeometricRegistration} or KPConv-style sparse convolutions in Predator~\cite{Huang2020PREDATOR:Overlap} use sophisticated architectures to properly capture local geometry. Similarly, LoGDesc~\cite{Slimani2024LoGDesc:Registration} propagates local geometric features globally through graph convolutions and attention mechanisms. PARE-Net~\cite{Yao2024PARE-Net:Registration} further addresses rotation sensitivity via position-aware rotation-equivariant framework. 

Another important evolution of learning approaches is that they perform correspondence finding in the learned feature space rather than directly on original coordinates. DCP~\cite{Wang2019DeepRegistration} and PRNet~\cite{Wang2019PRNet:Registration} optimize soft matching matrices, while RPMNet~\cite{Yew2020RPM-Net:Features} uses a differentiable Sinkhorn normalization layer~\cite{Sinkhorn1964AMatrices} to update a similarity matrix. Alternatively, Predator~\cite{Huang2020PREDATOR:Overlap} and GeoTransformer~\cite{Qin2022GeometricRegistration} predict overlaps between key points to focus on shared regions between two point clouds. Furthermore, some approaches incorporate an outlier filtering scheme to identify and suppress false correspondences. RPMNet~\cite{Yew2020RPM-Net:Features} learns an outlier rejection threshold via a parameter estimation network to exclude incorrect matches, while FastMAC~\cite{fastMAC_Zhang_2024_CVPR} utilizes graph signal processing to prioritize high-frequency nodes and allow for more efficient filtering of low-confidence outliers.

More recently, transformer-based models have been adopted, leveraging attention mechanisms to capture long-range dependencies and global geometric relationships. GeoTransformer~\cite{Qin2022GeometricRegistration} encodes distance and angle information, while REGTR~\cite{Yew2022REGTR:Transformers} combines self-attention and cross-attention mechanisms to improve correspondence prediction. Nevertheless, global self-attention can introduce ambiguity in low-overlap scenarios by correlating features across non-overlapping regions. To mitigate this issue, PEAL~\cite{Yu2023PEAL:Registration} integrates an overlap prior to categorize points into overlapping anchor and non-anchor regions, then employs a one-way attention module that restricts information flow from non-anchor to anchor points. Despite improved accuracy, these methods are computationally expensive.

In contrast, our method re-evaluates the need for engineered features and complex architectures, which limit networks to small point patches and introduce substantial computational costs. By expanding the receptive field, R3PM-Net reduces dependency on under-populated local neighborhoods and enables efficient real-time application.
\section{Method}
\label{sec:method}
\begin{figure*}
\centering
\includegraphics[width=1\textwidth]{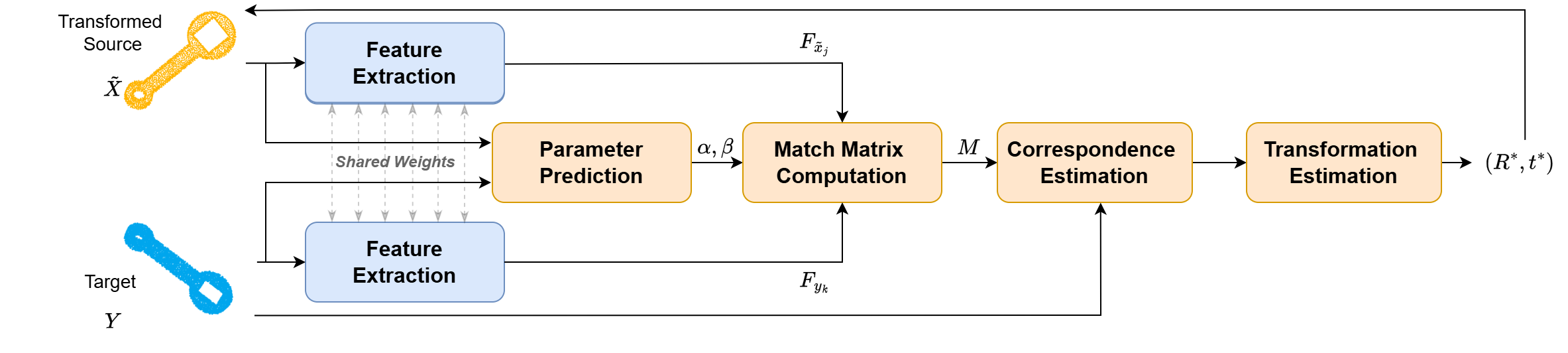} 
\caption{Overview of the R3PM-Net Architecture. Built upon~\cite{Yew2020RPM-Net:Features}, R3PM-Net employs an iterative Siamese framework for robust point cloud registration. The architecture includes four primary stages: (1) global-aware feature extraction employing shared MLPs to learn geometric similarities across a full receptive field; (2) correspondence estimation via a match matrix $M$ computed from feature distances; (3) outlier rejection driven by a parameter prediction module (4) transformation estimation using a differentiable SVD module to solve for the optimal rigid alignment $(R^*, t^*)$. The estimated transformation is applied back to the source point cloud for iterative refinement.}
\label{fig:r3pm-net}
\end{figure*}

The proposed R3PM-Net is a real-time deep learning architecture designed for the registration of sparse and imperfect industrial point clouds. R3PM-Net adopts RPMNet~\cite{Yew2020RPM-Net:Features} as baseline and reconsiders  its complex hybrid features to address the limitations of local geometric descriptors in noisy, imperfect real-world settings. 

As illustrated in Fig.~\ref{fig:r3pm-net}, R3PM-Net consists of four key components; Feature extraction, correspondence estimation, outlier rejection and transformation parameter estimation. The feature extraction component, the core contribution of R3PM-Net, employs a lightweight network with a global receptive field to process complete input point clouds directly and produce globally-aware features. Next, the correspondence estimation matches the points between the Source ($X$) and Target ($Y$) clouds by employing the extracted features. The outlier rejection module dynamically predicts thresholds to suppress false matches. Finally, a differentiable weighted Singular Value Decomposition (SVD) step~\cite{Arun1987Least-SquaresSets} predicts the optimal rigid transformation to align the clouds.



\subsection{Feature Extraction}

The feature extraction part of R3PM-Net is a light network with a global receptive field that directly maps raw 3D coordinates into a high-dimensional embedding space (inspired by~\cite{Qi2016PointNet:Segmentation}). This module defines a non-linear mapping function $\varphi$, which encodes each point $\mathbf{p} = (x, y, z)$ into a feature vector that captures the geometric structure around it as perceived by the network.

\begin{equation}
\varphi: \mathbb{R}^3 \to \mathbb{R}^D, \quad \text{where } D=1024,
\label{eq:mapping}
\end{equation}

\noindent To ensure that the source and target point clouds are processed consistently, R3PM-Net utilizes a Siamese architecture with shared weights. This ensures that the resulting features $\mathbf{F}_X$ and $\mathbf{F}_Y$ lie within a common embedding space, allowing for direct comparison based on Euclidean distance during the following matching stages:

\begin{equation}
\mathbf{X}, \mathbf{Y} \xrightarrow{\varphi} F_X, F_Y.
\end{equation}
The mapping $\varphi$ is implemented as a shared Multilayer Perceptron (MLP) of five linear layers with ReLU activations ($\sigma$), applied point-wise. This configuration allows for each point to be processed independently while undergoing the same learned transformation. Earlier layers capture fundamental cues, including local orientation and curvature, and deeper layers extract rich complex structural information. A final global max-pooling operation aggregates these point-wise features to incorporate the global geometric context. The efficacy of this module stems from the local similarity of feature vectors. By encoding both local characteristics and relative global positions, the network provides a low $L_2$ distance between the descriptors of the corresponding points. This enables R3PM-Net to establish robust matches even in the presence of sensor noise, and the sparsity typical of real-world industrial object-level scans.

\subsection{Correspondence Estimation}

Following the feature extraction step, the network predicts soft correspondences for the source and target point clouds. Instead of a binary assignment, a matching matrix $M \in [0,1]^{J \times K}$ is calculated, where each element $m_{jk}$ represents the probability that the point $x_j$ corresponds to the point $y_k$.

A deterministic annealing schedule~\cite{Yew2020RPM-Net:Features} is employed to avoid local minima, and the matrix is initialized based on the Euclidean distance of the learned features $F$:
\begin{equation}
m_{jk} \leftarrow e^{ -\beta \left( \| F_{x_j} - F_{y_k} \|_2^2 - \alpha \right) },
\label{eq:soft_matching}
\end{equation}
where $\alpha$ is a learned outlier threshold and $\beta$ is the annealing parameter controlling the ``sharpness" of the matching. Sinkhorn normalization~\cite{Sinkhorn1964AMatrices} is then applied to $\mathbf{M}$ to enforce bistochastic matrix constraints, ensuring that rows and columns sum to 1 (or handle outliers via slack variables).

\subsection{Outlier Rejection}

Unlike in simulated datasets created by applying random transformations to CAD models, real-world industrial point clouds often contain points without correspondence (outliers) due to varying sensor sources. In Eq. \ref{eq:soft_matching}, the parameter $\alpha$ acts as a decision boundary for incorrect outlier matches; if the feature distance $\| F_{x_j} - F_{y_k} \|_2^2 > \alpha$, the match probability is suppressed.

Instead of setting a static threshold, R3PM-Net follows~\cite{Yew2020RPM-Net:Features} and employs a PointNet module to dynamically predict $\alpha$ and $\beta$ at each iteration based on the current alignment state. This allows the network to be lenient in early iterations (soft matching) and strict in later iterations (hard matching), thereby, effectively filtering outliers as the registration improves.

\subsection{Transformation Parameter Estimation}
Once the match matrix $M$ is computed, the rigid transformation $\{R^*, t^*\}$ is estimated. For each source point $x_j$, a corresponding target coordinate $\hat{y}_j$ is computed as a weighted sum:
\begin{equation}
\hat{y}_j = \frac{1}{\sum_{k=1}^{K} m_{jk}} \sum_{k=1}^{K} m_{jk} \cdot y_k.
\label{eq:estimate corr}
\end{equation}
The optimal transformation is then solved in closed form using a weighted Singular Value Decomposition (SVD) module~\cite{Arun1987Least-SquaresSets}. Importantly, this SVD step is differentiable, allowing gradients to backpropagate through the transformation estimation to the feature extractor.

\subsection{Loss Function}

The network is trained end-to-end with a composite loss function: 
\begin{equation}
\mathcal{L}_{total} = \mathcal{L}_{reg} + \mathcal{L}_{geo \ align}.
\end{equation}
The primary term is the registration loss ($\mathcal{L}_{reg}$), defined as the $L_1$ distance of the source points transformed by the ground truth versus the predicted transformation:
\begin{equation}
\mathcal{L}_{reg} = \frac{1}{J} \sum_{j=1}^{J} \left\| (\mathbf{R}_{gt} \mathbf{x}_j + \mathbf{t}_{gt}) - (\mathbf{R}_{pred} \mathbf{x}_j + \mathbf{t}_{pred}) \right\|_1,
\end{equation}
where $\mathbf{R}_{gt}$ and $\mathbf{t}_{gt}$ are the ground-truth rotation matrix and translation vector, while $\mathbf{R}_{pred}$ and $\mathbf{t}_{pred}$ indicate the predicted rotation and translation.

To ensure correct matches, R3PM-Net incorporates a Geometric Alignment loss term, which measures the accuracy of matches. $\mathcal{L}_{geo \ align}$ is the $L_2$ distance between a point feature $F_{x_j}$ and its predicted counterpart; the weighted average of point features in the second set.
\begin{equation}
\mathcal{L}_{geo \ align} = \frac{1}{J} \sum_{j=1}^{J} \left\| F_{x_j} - \sum_{k=1}^{K} m_{jk} F_{y_k} \right\|_2^2
\label{eq:l_feat_dist}
\end{equation}

\subsection{Coarse-to-Fine Registration}
\label{sec:pipeline}

\begin{figure*}
\centering
\includegraphics[width=0.9\textwidth]{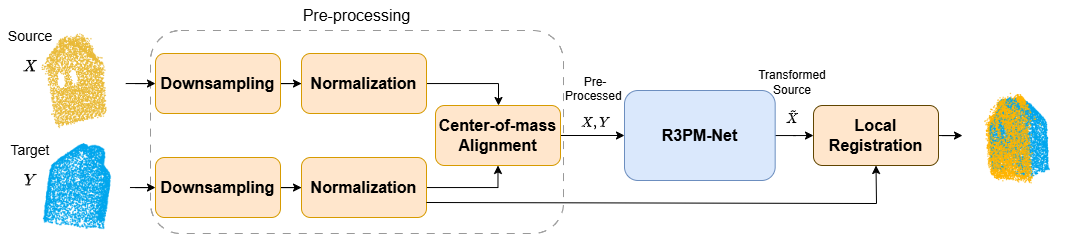} 
\caption{Coarse-to-Fine Registration approach to handle noisy, sparse scans. Initially, $X$ and $Y$ are pre-processed via downsampling, normalization and alignment. Large misalignments are then resolved through coarse global registration performed by R3PM-Net, followed by a final high-precision local refinement with GICP.}
\label{fig:pipeline}
\end{figure*}

To accommodate the industrial high-precision demands, R3PM-Net is integrated into a unified coarse-to-fine architecture (Fig.~\ref{fig:pipeline}). The process begins with the pre-processing of source ($X$) and target ($Y$) point clouds, which are uniformly downsampled, normalized, and centroid-aligned for numerical stability and memory efficiency. These clouds are then processed by R3PM-Net to provide a robust initial alignment. Finally, local refinement is performed via Generalized ICP (GICP)~\cite{Segal2010Generalized-ICP}. With the reliable global pose estimation of R3PM-Net, GICP avoids local minima and rapidly converges to a precise fit. This combination ensures reliable, real-time registration in challenging scenarios.

\section{Experiments}
\label{sec:experiments}

\begin{table*}
\centering
\setlength{\tabcolsep}{2pt}
\renewcommand{\arraystretch}{1.2}
\begin{tabular}{l | >{\centering\arraybackslash}p{2.3cm} >{\centering\arraybackslash}p{2.2cm} >{\centering\arraybackslash}p{2.2cm} >{\centering\arraybackslash}p{1.9cm} >{\centering\arraybackslash}p{3.1cm} >{\centering\arraybackslash}p{1.9cm}}
\toprule
\textbf{Method} & \textbf{RRE [$^{\circ}$] $\downarrow$}  & \textbf{RTE [cm]  $\downarrow$} & \textbf{CD [cm]  $\downarrow$}  & \textbf{Fitness $\uparrow$} & \textbf{Inlier RMSE [cm] $\downarrow$}  & \textbf{Time [s] $\downarrow$} \\
\midrule
RPMNet~\cite{Yew2020RPM-Net:Features} & 30.898 \scriptsize{$\pm$ 0.322} & \textbf{0.002 \scriptsize{$\pm$ 0.000}} & 0.153 \scriptsize{$\pm$ 0.001} & \underline{0.998 \scriptsize{$\pm$ 0.000}} & 0.094 \scriptsize{$\pm$ 0.001} &  \underline{0.021 \scriptsize{$\pm$ 0.000}} \\
Predator~\cite{Huang2020PREDATOR:Overlap} & 7.262 \scriptsize{$\pm$ 0.555} & 0.028 \scriptsize{$\pm$ 0.002} & \underline{0.045 \scriptsize{$\pm$ 0.002}} & \textbf{1.000 \scriptsize{$\pm$ 0.000}} & \underline{0.026 \scriptsize{$\pm$ 0.001}} & 0.071 \scriptsize{$\pm$ 0.001} \\
GeoTransformer~\cite{Qin2022GeometricRegistration} & 50.357 \scriptsize{$\pm$ 1.238} & 0.215 \scriptsize{$\pm$ 0.003} & 0.255 \scriptsize{$\pm$ 0.005} & 0.921 \scriptsize{$\pm$ 0.001} & 0.101 \scriptsize{$\pm$ 0.002} & 0.065 \scriptsize{$\pm$ 0.000} \\
RegTR~\cite{Yew2022REGTR:Transformers} & \textbf{1.712 \scriptsize{$\pm$ 0.061}} & \underline{0.007 \scriptsize{$\pm$ 0.000}} & \textbf{0.017 \scriptsize{$\pm$ 0.000}} & \textbf{1.000 \scriptsize{$\pm$ 0.000}} & \textbf{0.009 \scriptsize{$\pm$ 0.000}} & 0.045 \scriptsize{$\pm$ 0.000} \\
LoGDesc~\cite{Slimani2024LoGDesc:Registration} & 42.762 \scriptsize{$\pm$ 1.254} & 0.158 \scriptsize{$\pm$ 0.003} & 0.183 \scriptsize{$\pm$ 0.004} & 0.978 \scriptsize{$\pm$ 0.001} & 0.097 \scriptsize{$\pm$ 0.002} & 0.075 \scriptsize{$\pm$ 0.000} \\
\midrule
R3PM-Net (ours) &  \underline{5.198 \scriptsize{$\pm$ 0.067}} &  0.010 \scriptsize{$\pm$ 0.000} &  0.052 \scriptsize{$\pm$ 0.000} & \textbf{1.000 \scriptsize{$\pm$ 0.000}} &  0.029 \scriptsize{$\pm$ 0.000} & \textbf{0.007 \scriptsize{$\pm$ 0.000}} \\
\bottomrule
\end{tabular}
\caption{Quantitative comparison on ModelNet40. Performance is reported in terms of RRE, RTE, Chamfer distance (CD), fitness (↑ higher is better), inlier RMSE, and inference time (↓ lower is better). R3PM-Net achieves the fastest inference while maintaining competitive accuracy. \textbf{Best} and \underline{second-best} results are highlighted.}
\label{tab:Modelnet40}
\end{table*}

\begin{table*}
\centering
\setlength{\tabcolsep}{2pt}
\renewcommand{\arraystretch}{1.2}
\begin{tabular}{l | >{\centering\arraybackslash}p{2.3cm} >{\centering\arraybackslash}p{2.2cm} >{\centering\arraybackslash}p{2.2cm} >{\centering\arraybackslash}p{1.9cm} >{\centering\arraybackslash}p{3.1cm} >{\centering\arraybackslash}p{1.9cm}}
\toprule
\textbf{Method} & \textbf{RRE [$^{\circ}$] $\downarrow$}  & \textbf{RTE [cm]  $\downarrow$} & \textbf{CD [cm]  $\downarrow$}  & \textbf{Fitness $\uparrow$} & \textbf{Inlier RMSE [cm] $\downarrow$}  & \textbf{Time [s] $\downarrow$} \\
\midrule
RPMNet~\cite{Yew2020RPM-Net:Features} & 32.217 \scriptsize{$\pm$ 0.844} & \textbf{0.002 \scriptsize{$\pm$ 0.000}} & 0.160 \scriptsize{$\pm$ 0.004} & \underline{0.997 \scriptsize{$\pm$ 0.002}} & 0.098 \scriptsize{$\pm$ 0.001} &  0.021 \scriptsize{$\pm$ 0.000} \\
Predator~\cite{Huang2020PREDATOR:Overlap} & 16.448 \scriptsize{$\pm$ 1.548} & 0.044 \scriptsize{$\pm$ 0.003} & 0.072 \scriptsize{$\pm$ 0.002} & \textbf{1.000 \scriptsize{$\pm$ 0.000}} & 0.042 \scriptsize{$\pm$ 0.001} & 0.071 \scriptsize{$\pm$ 0.001} \\
GeoTrans.~\cite{Qin2022GeometricRegistration} & 45.582 \scriptsize{$\pm$ 0.549} & 0.183 \scriptsize{$\pm$ 0.005} & 0.297 \scriptsize{$\pm$ 0.007} & 0.906 \scriptsize{$\pm$ 0.004} & 0.111 \scriptsize{$\pm$ 0.002} & 0.065 \scriptsize{$\pm$ 0.000} \\
RegTR~\cite{Yew2022REGTR:Transformers} & \textbf{1.311 \scriptsize{$\pm$ 0.032}} & \underline{0.004 \scriptsize{$\pm$ 0.000}} & \textbf{0.023 \scriptsize{$\pm$ 0.000}} & \textbf{1.000 \scriptsize{$\pm$ 0.000}}  & \textbf{0.012 \scriptsize{$\pm$ 0.000}} & 0.045 \scriptsize{$\pm$ 0.000} \\
LoGDesc~\cite{Slimani2024LoGDesc:Registration} & 121.224 \scriptsize{$\pm$ 10.396} & 0.773 \scriptsize{$\pm$ 0.060} & 0.692 \scriptsize{$\pm$ 0.022} & 0.718 \scriptsize{$\pm$ 0.010} & 0.224 \scriptsize{$\pm$ 0.002} & 0.075 \scriptsize{$\pm$ 0.000} \\
\midrule
R3PM-Net (ours) &  \underline{5.451 \scriptsize{$\pm$ 0.287}} &  0.006 \scriptsize{$\pm$ 0.001} &  \underline{0.054 \scriptsize{$\pm$ 0.002}} & \textbf{1.000 \scriptsize{$\pm$ 0.000}} &  \underline{0.030 \scriptsize{$\pm$ 0.001}} & \textbf{0.006 \scriptsize{$\pm$ 0.000}} \\
\bottomrule
\end{tabular}
\caption{Quantitative comparison on Sioux-Cranfield dataset. R3PM-Net offers the best trade-off between real-time execution and competitive accuracy; it remains highly competitive, exceeding the performance of recent methods on most metrics while operating at an inference latency over $6.5\times$ faster than RegTR. \textbf{Best} and \underline{Second-best} results are highlighted.}
\label{tab:sioux-cranfield}
\end{table*}

\subsection{Datasets}\label{subsec:datasets} 
\textbf{ModelNet40}~\cite{Wu20143DShapes} is a collection of synthetic 3D CAD models from 40 object categories. The official dataset split includes 9,843 samples (80\%) for training and 2,468 (20\%) for testing. The experiments in this paper are done on the testing set only to evaluate the performance on clean and ideal data.

\noindent\textbf{Sioux-Cranfield}, proposed by this paper, is a diverse collection of 13 objects designed to evaluate model robustness across varying data qualities. The dataset contains 4 computer-aided design (CAD) models generated via photogrammetric reconstruction~\cite{Griwodz2021AliceVisionMeshroom}, 3 synthetic CAD models, and 6 pristine geometries from the Cranfield Benchmark~\cite{Collins1985DEVELOPMENTSYSTEMS., Denayer2024ComparisonObjects}. This combination allows for a comprehensive evaluation of performance on both high-quality synthetic meshes and realistically imperfect reconstructions.

\noindent\textbf{Sioux-Scans} is another dataset introduced in this work that addresses the real-world challenge of registering physical scans to digital models. The targets are CAD models of seven small objects (shared with Sioux-Cranfield), while the sources are raw event-camera scans of the corresponding objects acquired via the custom Quality Control \textit{Sioux 3DoP} setup~\cite{2023OptimizationAligners}. To generate these scans, the setup utilizes a laser beam and an event-based camera to produce accurate point clouds from moving or handheld objects. Unlike traditional frame-based sensors, this camera captures discrete brightness changes as the laser sweeps across the surface, resulting in highly precise point clouds. However, these data represent a substantially more challenging setting than synthetic benchmarks, as they reflect inevitable deficiencies, such as sparsity, noise, and occlusions, rarely present in ideal simulated datasets. These artifacts stem from sensor noise, lighting sensitivity, and viewpoint-dependent gaps, particularly on sharp edges or reflective surfaces.

Following~\cite{Wang2019DeepRegistration} and~\cite{Yew2020RPM-Net:Features}, all point clouds are downsampled and normalized to a unit sphere. Except for the \textit{Sioux-Scans} data, the source-target pairs are simulated by applying random rotations in the range of $[0^{\circ}, 45^{\circ}]$ and translations in $[-0.5, 0.5]$ cm along each axis. These transformations are applied to a copy of each point cloud to form the source point cloud $X$, while the goal is to register these point clouds to the untransformed target $Y$. Visualizations and dataset details are provided in the supplementary material (section~\ref{sup:datasets})

\subsection{Evaluation Metrics}\label{subsec:metrics}
Following previous works~\cite{Wang2019DeepRegistration, Wang2019PRNet:Registration, Yew2020RPM-Net:Features, Huang2020PREDATOR:Overlap}, performance on object-level point cloud registration is evaluated by relative rotation error (RRE), relative translation error (RTE) and Chamfer distance (CD). In addition, fitness (measures overlapping areas of two point clouds and is defined as the ratio of the number of inlier correspondences to the total number of target points), inlier RMSE~\cite{Zhou2018Open3D:Processing}, and inference time metrics are reported. 
For a comprehensive evaluation, these metrics should be considered jointly. While they provide a reliable quantitative assessment, visual inspection is required to verify accurate registration, particularly on Sioux-Scans. Additional details and the mathematical definitions of all metrics are provided in the supplementary material (section~\ref{sup:metrics}).

\subsection{Implementation Details}
 R3PM-Net employs two PointNet networks~\cite{Qi2016PointNet:Segmentation} pre-trained on ModelNet40~\cite{Wu20143DShapes}, which share weights, as feature extractions. This architecture is not end-to-end trained further. Experiments are conducted with the official pre-trained models (also on ModelNet40) provided by the authors and the implementation of RPMNet by~\cite{Sarode2020Learning3d}. All inference benchmarks are conducted on a single NVIDIA H100 GPU.

%

\subsection{Experimental Results}\label{subsec:results}
R3PM-Net is compared against state-of-the-art methods spanning a diverse set of PCR paradigms, including RPMNet~\cite{Yew2020RPM-Net:Features} (\textit{point-based correspondence}), Predator~\cite{Huang2020PREDATOR:Overlap} and GeoTransformer~\cite{Qin2022GeometricRegistration} (\textit{voxel-based overlap detection}), REGTR~\cite{Yew2022REGTR:Transformers} (\textit{transformer-based correspondence search}), and LoGDesc~\cite{Slimani2024LoGDesc:Registration} (\textit{hybrid method combining graph convolutions and attention mechanisms}).

To ensure statistical stability, reproducibility, and a fair comparison across all models, each evaluation is repeated over seven independent runs with different random seeds. The results in Tables~\ref{tab:Modelnet40},~\ref{tab:sioux-cranfield}, and~\ref{tab:unified_results} represent the mean and standard deviation of these iterations. 

\begin{table*}
\centering
\small
\setlength{\tabcolsep}{2pt} 
\renewcommand{\arraystretch}{1.3}
\begin{tabular}{@{}l|ccc|ccc|ccc|ccc|c|c@{}}
\toprule
\multirow{2}{*}{\textbf{Method}} & \multicolumn{3}{c|}{\textbf{Lime}} & \multicolumn{3}{c|}{\textbf{Cube}} & \multicolumn{3}{c|}{\textbf{House}} & \multicolumn{3}{c|}{\textbf{Teeth}} & \textbf{Success} & \textbf{Time} \\
 & CD & Fit. & RMSE & CD & Fit. & RMSE & CD & Fit. & RMSE & CD & Fit. & RMSE & \textbf{Rate} & \textbf{(s)} \\
\midrule
RPMNet~\cite{Yew2020RPM-Net:Features} & 0.270 & 1.000 & 0.047 & 0.290 & 1.000 & 0.023 & - & - & - & - & - & - & 28.6\% & 0.042 \\
Predator~\cite{Huang2020PREDATOR:Overlap} & 0.270 & 1.000 & 0.048 & 0.289 & 1.000 & 0.023 & - & - & - & - & - & - & 28.6\% & \textbf{0.038} \\
GeoTrans.~\cite{Qin2022GeometricRegistration} & 0.260 & 1.000 & 0.041 & 0.295 & 1.000 & 0.024 & - & - & - & - & - & - & 28.6\% & 0.042 \\
RegTR~\cite{Yew2022REGTR:Transformers} & 0.270 & 1.000 & 0.047 & 0.292 & 1.000 & 0.023 & - & - & - & - & - & - & 28.6\% & \textbf{0.038} \\
LoGDesc~\cite{Slimani2024LoGDesc:Registration} & - & - & - & 0.292 & 1.000 & 0.024 & 0.222 & 1.000 & 0.052 & - & - & - & 28.6\% & 0.043 \\
\midrule
R3PM-Net (ours)& - & - & - & 0.510 & 0.912 & 0.102 & - & - & - & 0.178 & 1.000 & 0.047 & 28.6\% & \underline{0.041} \\
\bottomrule
\end{tabular}
\caption{Registration performance for successful cases of Sioux-Scans dataset (visually verified). '-' indicates that the method did not solve the object case. Success Rate is the ratio of successful trials to the seven test objects. Runtime reflects the average time (in seconds) to attempt all seven cases. Please refer to the Supplementary material (section~\ref{sup:full_results}) for the full table including all failed cases.}
\label{tab:unified_results}
\end{table*}

\noindent\textbf{ModelNet40}.
As shown in Table~\ref{tab:Modelnet40}, R3PM-Net demonstrates superior efficiency while achieving highly competitive precision. It achieves the state-of-the-art rotation error, following RegTR closely, lower than other models, while maintaining a perfect Fitness score of $1.000$. More significantly, R3PM-Net demonstrates strong computational efficiency. Operating at $0.007$ s, which is $6.5\times$ faster than RegTR and an order of magnitude faster than LoGDesc, offering the best trade-off between registration precision and real-time latency. A detailed comparison of model parameters and throughput is provided in Table~\ref{tab:parameter_comparison}, highlighting R3PM-Net’s high efficiency.


\noindent\textbf{Sioux-Cranfield}.
To assess the generalizability and robustness of the compared methods, all models are evaluated on the proposed Sioux-Cranfield dataset. Table~\ref{tab:sioux-cranfield} indicates that R3PM-Net demonstrates stable performance, maintaining a perfect Fitness score of $1.000$ and outperforming Predator, GeoTransformer, and LoGDesc across nearly all metrics. R3PM-Net achieves RRE and RTE results comparable to the state-of-the-art RegTR, yet with a decisive efficiency advantage. R3PM-Net operates at an inference speed over $6.5\times$ faster than RegTR, regardless of dataset complexity. This consistent performance gap confirms that high-precision registration does not necessitate the heavy computational load typical of current transformers, proving R3PM-Net as a solution for real-time, real-world applications where latency is indispensable.


\noindent\textbf{Sioux-Scans}
To evaluate the practical applicability of R3PM-Net, experiments are conducted on the novel Sioux-Scans dataset, which addresses the challenge of registering event-camera scans to digital CAD models. These scans exhibit inherent sparsity, noise, and occlusions, making the registration task significantly more difficult than synthetic benchmarks. Furthermore, since absolute ground-truth transformations are unavailable, evaluation relies on metrics that do not require ground-truth information (Chamfer distance, fitness and inlier RMSE), complemented by visual inspection of the alignment quality.
As summarized in Table~\ref{tab:unified_results}, R3PM-Net matches the $28.6\%$ success rate of the baselines through a minimalist feature-extraction approach, in contrast to the more complex backbones employed by other methods. While all models solve the less challenging symmetrical cases, R3PM-Net successfully registers objects with complex geometries, such as the ``teeth" model, where all other approaches fail. Despite the increased complexity of these data, R3PM-Net maintains a competitive average runtime of $41$ ms, comparable to the fastest baseline methods, while yielding success on more difficult geometries.

Although R3PM-Net solves a number of edge cases, achieving complete success remains challenging due to the inherent noise, outliers and occlusions in event-camera scans, particularly in feature-sparse objects such as ``Lego", where overlapping regions are insufficient. While baseline methods struggle with the non-convex complexities of models such as ``teeth", R3PM-Net preserves geometric details that traditional descriptors often smooth out. This allows R3PM-Net to sustain robust correspondences even in presence of artifacts and low point density.



\begin{figure}[htbp]
\centering
\includegraphics[width=0.4\textwidth]{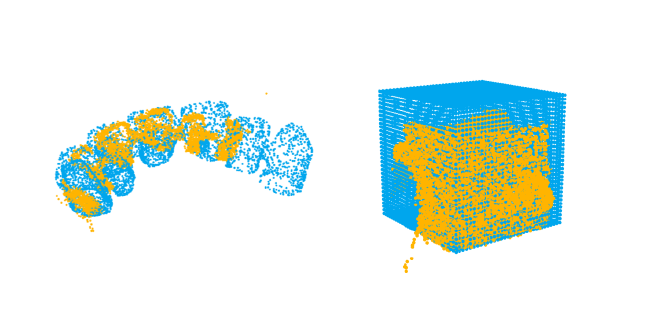} 
\vspace{-6mm}
\caption{Qualitative registration results of R3PM-Net on real-world event-camera data. It successfully aligns the ``teeth" and ``cube" models in under $50$ ms.}
\label{fig:qualitative-results}
\end{figure}

\begin{table} [h]
\centering
\small 
\setlength{\tabcolsep}{3pt}
\begin{tabular}{l|cc} 
\toprule
\textbf{Model} & \textbf{Total Params. [M] $\downarrow$} & \textbf{Throughput [fps]} \\
\midrule
RPMNet                   & \textbf{0.91} & \underline{48} \\
Predator                 & 22.57         & 14             \\
GeoTrans.                & 5.21          & 15            \\
RegTR                    & 11.49         & 22             \\
LoGDesc                  & 4.71          & 13             \\
\midrule
\textbf{R3PM-Net (ours)} & \underline{0.96} & \textbf{167} \\
\bottomrule
\end{tabular}
\caption{Model complexity and throughput comparison. Throughput is the number of point cloud pairs registered per second. Despite comparable performance, R3PM-Net reduces total parameters by over $90\%$ compared to RegTR.}
\label{tab:parameter_comparison}
\end{table}

\subsection{Ablation Studies}\label{subsec:ablations}

Ablation studies are performed to justify the choice of input representation of R3PM-Net architecture and to demonstrate that fine-tuning on the proposed Sioux-Cranfield dataset enhances performance across all evaluated benchmarks, including standard public datasets. To this end, R3PM-Net (FT) is fine-tuned end-to-end for 50 epochs on a subset of the Sioux-Cranfield dataset, excluded from evaluation to prevent data leakage. Optimization is performed via Adam ($LR=0.001$) on an NVIDIA RTX A5000 GPU. Additionally, the sensitivity of the fine-tuned R3PM-Net to the composition of the fine-tuning data subsample is studied.

\noindent\textbf{Input Features}.
This experiment examines the impact of incorporating hand-crafted local features (PC feat.), such as surface normals and neighborhood surfaces defined by fixed or flexible radii. In the flexible radius setting, the radius is dynamically selected as the maximum distance from each point to its nearest neighbor, ensuring the inclusion of a sufficient number of points.

While such features are commonly employed to enhance local geometric awareness in dense synthetic datasets, the results in Table~\ref{tab:input_ablation} show that these features limit the field-of-view to local features, and significantly degrade performance when applied to datasets containing sparse, imperfect point clouds. Specifically the inclusion of surface normals decreases performance due to the instability of normal estimation in non-synthetic datasets, which introduces additional noise into the feature space, thereby, hindering learning. In particular, the combination of normals and fixed radii increases the RRE to $31.862$$^\circ$ and triples the runtime to $0.021$s. These hand-crafted constraints restrict the network's effective field-of-view, limiting the ability to capture global contexts. In contrast, the Direct PC configuration, i.e, where the network receives complete pure point clouds in its receptive field, achieves superior performance across nearly all metrics, supporting the design choice of directly processing point clouds to ensure a real-time architecture that maintains a sufficient field-of-view.

\noindent\textbf{Effectiveness of Fine-Tuning}.
Fine-tuning on a subset of the proposed Sioux-Cranfield dataset acts as a robust regularizer, considerably  enhancing performance across all benchmarks. As shown in Table~\ref{tab:fine_tuning_ablate}, the FT variant reduces rotation error by over $50\%$ on both ModelNet40 and Sioux-Cranfield. More significantly, as indicated in Table~\ref{tab:ablation_subsets}, this fine-tuning process nearly doubles the success rate in high-sparsity Sioux-Scans ($28.6\%$ to $42.9\%$). This proves that learning from imperfect, challenging data provides a superior training signal for handling noise and sparsity compared to purely synthetic data, all while maintaining a $6.5\times$ speed advantage over complex backbones.

\noindent\textbf{Fine-Tuning Subsets}.
Using the introduced Sioux-Cranfield dataset, multiple subsets are generated to analyze the impact of fine-tuning data composition on R3PM-Net performance on event-camera scans. Table~\ref{tab:ablation_subsets} indicates that fine-tuning on subsets that span diverse geometric structures and symmetrical shapes, such as ``teeth", ``lime", ``cube", ``Lego", ``round-peg", ``separator", and ``shoe", yields the highest success rate of $42.86\%$. Notably, the model fine-tuned on the latter subset successfully registers the complex ``teeth" object without having the CAD model in its fine-tuning data. This proves that the proposed network learns fundamental geometric primitives (e.g., local curvature or edge patterns) rather than memorizing object-specific shapes.
The model fine-tuned on the first subset (``teeth", ``lime", ``cube" and ``Lego") is the only configuration that successfully registers the challenging ``house" object, suggesting that the inclusion of ``Lego", which has many sharp 90-degree angles and planar surfaces, can be the key to helping the model understand the shape of the ``house" scan.
Conversely, fine-tuning on subsets composed of similar or symmetric objects (e.g., ``B-t-Plate", ``elephant", ``house", ``round-peg" and ``shoe") leads to feature interference and worse results. Similarly, fine-tuning on the full dataset results in overfitting, reducing generalization, and overall performance.

\begin{table}
\centering
\footnotesize 
\setlength{\tabcolsep}{3pt} 
\renewcommand{\arraystretch}{1.3}
\begin{tabular}{>{\raggedright\arraybackslash}m{1.6cm}|cc|ccccc}
\toprule
\textbf{Input Rep.} & \textbf{Norm.} & \textbf{Rad.} & \textbf{RRE $\downarrow$} & \textbf{RTE $\downarrow$} & \textbf{CD $\downarrow$} & \textbf{Fit. $\uparrow$} & \textbf{Time} \\
\midrule
PC Feat.1 & \checkmark & Fixed & 31.86  & \textbf{0.00} & 0.16 & \textbf{1.00} & 0.021 \\
PC Feat.2 & -- & Fixed & 9.35  & 0.01 & 0.07 & \textbf{1.00} & \textbf{0.006} \\
PC Feat.3 & \checkmark & Flex & 13.65  & 0.01 & 0.08 & 0.99 & \textbf{0.006} \\
PC Feat.4 & -- & Flex & 12.90 & 0.01 & 0.08 & 0.99 & \textbf{0.006} \\
\midrule
\textbf{Direct PC R3PM-Net} & -- & -- & \textbf{2.01}  & \textbf{0.00} & \textbf{0.02} & \textbf{1.00} & \textbf{0.006} \\
\bottomrule
\end{tabular}
\caption{Ablation study on input features. Hand-crafted local features are compared to R3PM-Net's direct point cloud (PC) approach on the Sioux-Cranfield dataset. \textbf{Bold} indicates the best results.}
\label{tab:input_ablation}
\end{table}

\begin{table}[htbp]
\centering
\footnotesize
\setlength{\tabcolsep}{3pt} 
\renewcommand{\arraystretch}{1.4}
\begin{tabular}{p{1.3cm} | l | c c c c c} 
\toprule
\textbf{Dataset} & \textbf{Method} & \textbf{RRE $\downarrow$} & \textbf{RTE $\downarrow$} & \textbf{CD $\downarrow$} & \textbf{Fit. $\uparrow$} & \textbf{Time} \\
\midrule
\multirow{2}{1.3cm}{Model-Net40} 
& R3PM-Net & 5.198 & 0.010 & 0.052 & 0.029 & \textbf{0.007} \\
& R3PM-Net (FT) & \textbf{1.963} & \textbf{0.003} & \textbf{0.025} & \textbf{0.014} & \textbf{0.007} \\
\midrule
\multirow{2}{1.3cm}{Sioux-Cranfield} 
& R3PM-Net & 5.451 & 0.006 & 0.054 & 0.030 & \textbf{0.006} \\
& R3PM-Net (FT) & \textbf{2.297} & \textbf{0.002} & \textbf{0.033} & \textbf{0.018} & \textbf{0.006} \\
\bottomrule
\end{tabular}
\caption{Ablation study of cross-domain fine-tuning (FT). Fine-tuning on a subset of the proposed Sioux-Cranfield dataset improves performance across both dataset benchmarks. \textbf{Bold} indicates best results.}
\label{tab:fine_tuning_ablate}
\end{table}

\begin{table}[htbp]
\centering
\small 
\setlength{\tabcolsep}{4pt}
\begin{tabularx}{\columnwidth}{l|>{\RaggedRight\arraybackslash}X|>{\RaggedRight\arraybackslash}X|c}
\toprule
\textbf{Method} & \textbf{FT Subset} & \textbf{Success Cases} & \textbf{Rate (\%)} \\
\midrule
R3PM-Net & --- & teeth, cube & 28.6 \\
\midrule
R3PM-Net (FT)* & teeth, lime, cube, lego & teeth, lime, house & \textbf{42.9} \\
\midrule
\multirow{7}{*}{R3PM-Net (FT)} & rd-peg, sep, shoe & teeth, lime, cube & \textbf{42.9} \\
\cline{2-4}
 & Plate, eleph., house & cube & 14.3 \\
\cline{2-4}
 & Plate, eleph., house, rd-peg & teeth, cube & 28.6 \\
\cline{2-4}
 & rd-peg, sep, shoe, lego & teeth & 14.3 \\
\cline{2-4}
 & All 13 CADs & teeth, cube & 28.6 \\
\bottomrule
\end{tabularx}
\caption{Ablation study on fine-tuning subsets of Sioux-Cranfield dataset evaluated on Sioux-Scans.* denotes the model studied in the effectiveness of fine-tuning experiment shown in Table~\ref{tab:fine_tuning_ablate}.}
\label{tab:ablation_subsets}
\end{table}

\section{Conclusion}
\label{sec:conclusion}
This paper introduces R3PM-Net, a robust real-time point matching network designed to bridge the gap between synthetic benchmarks and real-world object-level industrial data. By choosing an expanded receptive field over complex architectures or hybrid features, R3PM-Net enables efficient registration of sparse, noisy, and occluded point clouds. Extensive evaluations demonstrate that R3PM-Net competes with significantly more sophisticated state-of-the-art models on synthetic and real-world datasets, such as ModelNet40 and the introduced \textit{Sioux-Cranfield} and \textit{Sioux-Scans} datasets, while operating at a fraction of their computational cost.

This paper highlights the gap in registration of point clouds in non-ideal real-life industrial settings. Since the existing methods are mostly focused on synthetic datasets that do not reflect the complications of real-world data, further research is needed to develop models that are robust to the variability and imperfections in real scans. Improving generalization and accuracy across diverse shapes, densities, structures, and perturbation levels remains a challenge in point cloud registration.

\section*{Acknowledgment}
This work is supported by Sioux Technologies and the ELEVATION Xecs 2023022 project.
{

}

\clearpage
\setcounter{page}{1}
\maketitlesupplementary

\renewcommand{\thesection}{A.\arabic{section}}
\setcounter{section}{0}
\renewcommand{\thesubsection}{\thesection.\arabic{subsection}}

\section{Datasets} \label{sup:datasets}

\textbf{ModelNet40 Dataset.}~\cite{Wu20143DShapes} is a collection of synthetic 3D CAD models from 40 object categories. To generate point clouds, 2,000 points are randomly sampled from the mesh surfaces of each model and normalized to a unit sphere.

\begin{figure*}[htbp]
\centering
\includegraphics[width=0.9\textwidth]{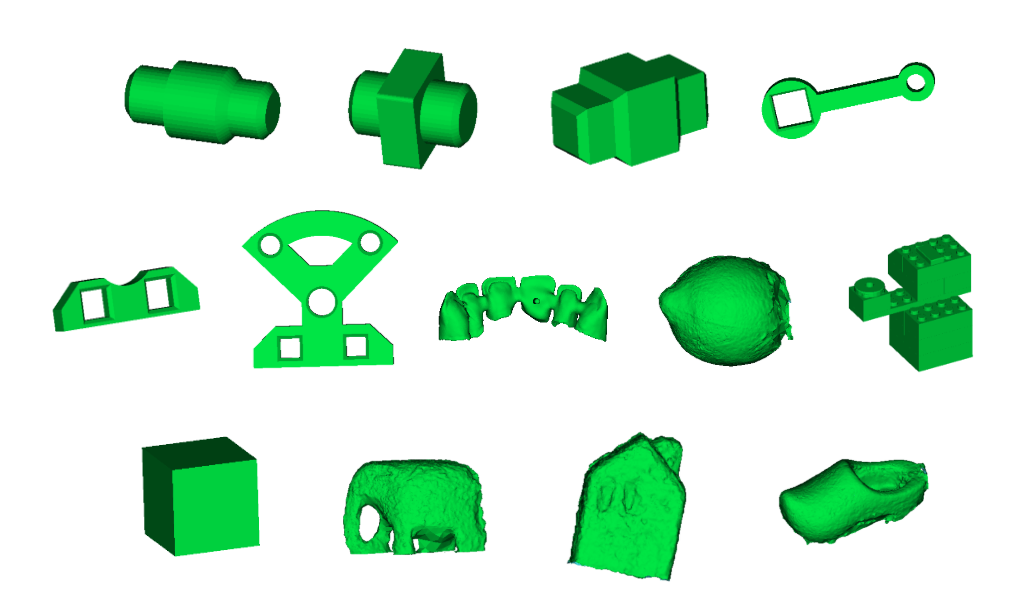} 
\caption{CAD models of the Sioux-Cranfield dataset. The first six belong to the Cranfield Assembly benchmark~\cite{Collins1985DEVELOPMENTSYSTEMS.} and the rest are contributions of this paper (Sioux dataset).}
\label{fig:sioux-cranfield}
\end{figure*}

\noindent\textbf{The Sioux-Cranfield Dataset.} is a diverse collection of 13 objects designed to evaluate model robustness across varying data qualities. The CAD models are presented in Fig.~\ref{fig:sioux-cranfield} while Table~\ref{tab:composition} provides a structured breakdown of the composition of this dataset.

\begin{figure}
\centering
\includegraphics[width=0.4\textwidth]{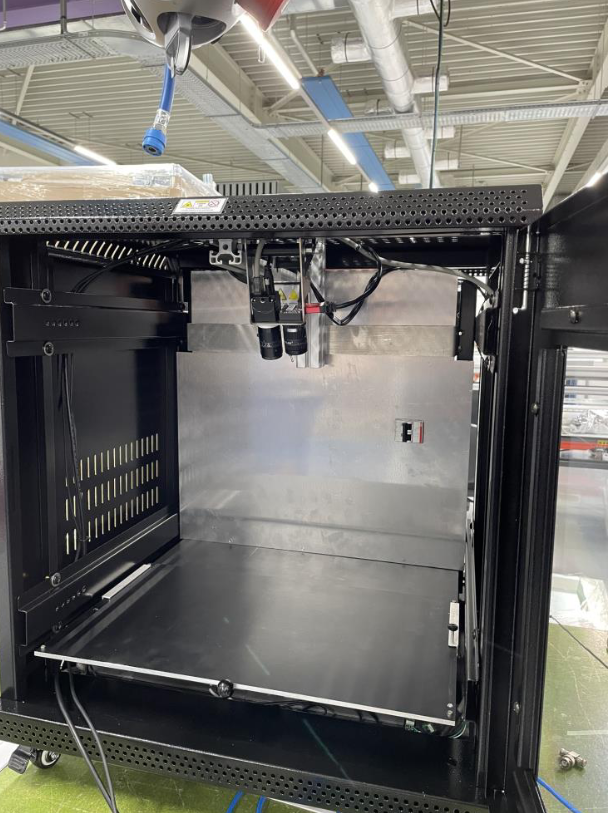} 
\caption{3D Quality Control Setup developed by Sioux Technologies~\cite{2023OptimizationAligners}. This setup leverages event-camera technology to provide fast, high-fidelity 3D scanning suitable for in-line real-time quality assurance for specialized products. It aims to deliver optimized process control in applications including 3D metal printing, integrated electronics in molded parts, and dental prosthetics.}
\label{sup:sioux_setup}
\end{figure}

\begin{figure*}[htbp]
\centering
\vspace{-10mm}
\includegraphics[width=0.9\textwidth]{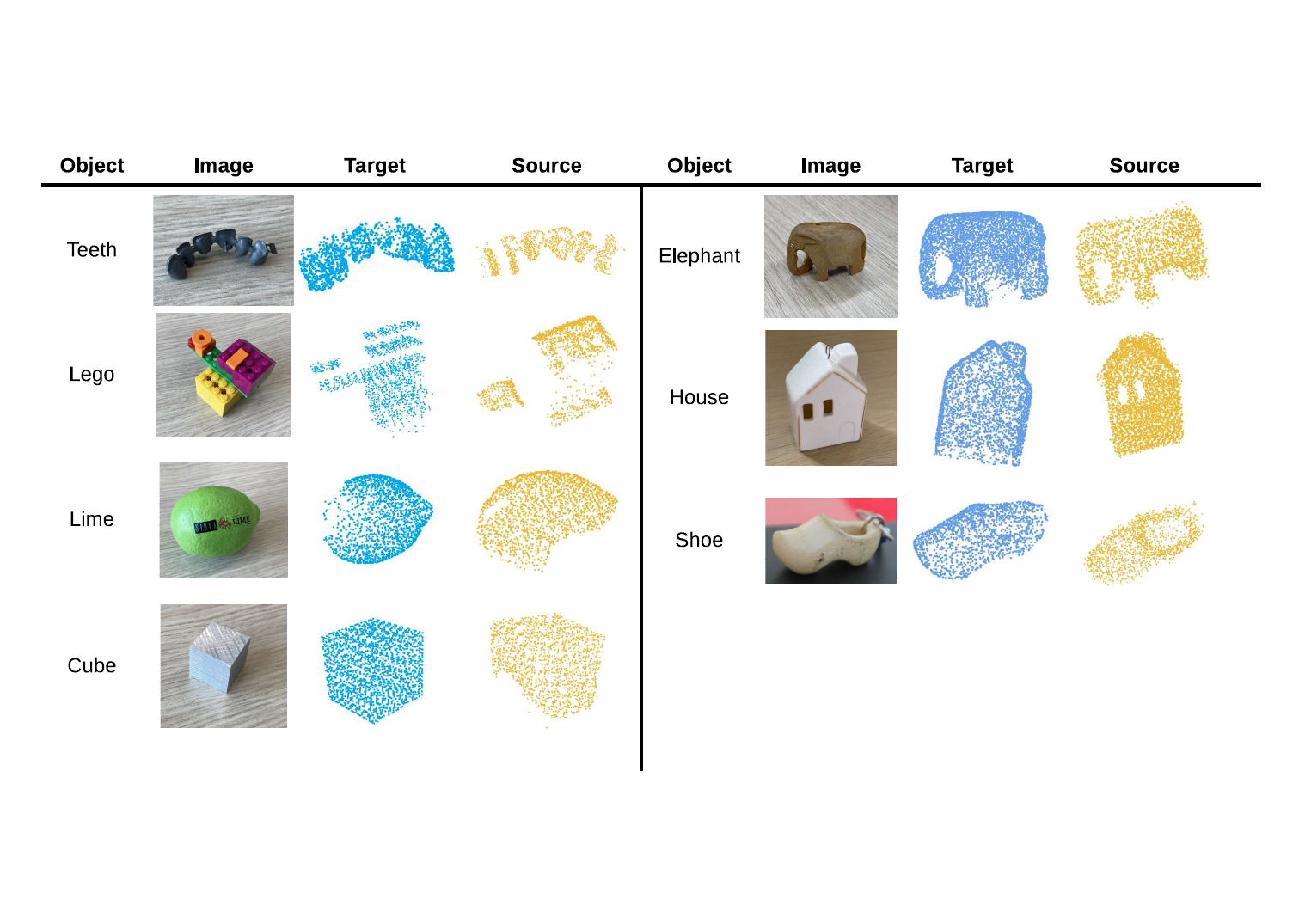} 
\vspace{-15mm}
\caption{Sioux-Scans point cloud data overview. Target (blue) and Source (yellow) point clouds for seven distinct objects.}
\label{fig:real-world-data}
\end{figure*}

\noindent\textbf{Sioux-Scan Data.} represents the core challenge of this work: the registration of raw event-camera scans against digital models. Unlike the simulated datasets described above, these pairs exhibit a genuine domain gap. The \textbf{Target} point clouds were derived from 3D CAD models of seven small objects (also present in the Sioux-Cranfield dataset). The \textbf{Source} point clouds were acquired by scanning the physical objects using an event camera within a 3D Quality Control Setup~\cite{2023OptimizationAligners} developed by Sioux Technologies (Fig.~\ref{sup:sioux_setup}). Before processing, gross outliers were filtered using \textit{CloudCompare}~\cite{girardeau2016cloudcompare}. As Fig.~\ref{fig:real-world-data} indicates, these scans exhibit severe unavoidable flaws which are not present in synthetic benchmarks, including high sparsity, sensor noise, and significant occlusions, particularly on object undersides and sharp edges hidden from the camera's field of view.

\begin{table}[htbp]
\centering
\small 
\setlength{\tabcolsep}{12pt} 
\begin{tabular}{llc} 
\toprule
\textbf{Category} & \textbf{Source Type} & \textbf{Qty} \\ \midrule
Sioux (Reconstructed) & Photogram. &  4 \\
Sioux (Synthetic)     & CAD Models &  3 \\
Cranfield~\cite{Collins1985DEVELOPMENTSYSTEMS.} & Pristine &  6 \\ \midrule
\textbf{Total}        & ---        & \textbf{13} \\ \bottomrule
\end{tabular}
\caption{Composition of the Sioux-Cranfield Dataset. Sioux (Reconstructed) and Sioux (Synthetic) are also used in Sioux-Scans dataset to produce Target point clouds.}
\label{tab:composition}
\end{table}

\section{Evaluation Metrics} \label{sup:metrics}


\begin{table*}[t]
\centering
\renewcommand{\arraystretch}{1.5}
\setlength{\tabcolsep}{6pt} 
\small 

\begin{tabularx}{\textwidth}{l l X X c}
\toprule
\textbf{Metric} & \textbf{Type} & \textbf{Pros} & \textbf{Cons} & \textbf{GT-Free} \\
\midrule
RRE & Geodesic dist. & Scale-invariant & Less intuitive, not always objective & \ding{55} \\
RTE & Euclidean dist. & Scale-invariant & Less intuitive, not always objective & \ding{55} \\
Chamfer Dist. & Euclidean dist. & Suitable for symmetric shapes & Scale-variant & \ding{51} \\
Fitness & Overlap ratio & Intuitive & Scale-variant & \ding{51} \\
Inlier RMSE & Euclidean dist. & Practical & Scale-variant, can be misleading with poor alignment & \ding{51} \\
Time & Seconds & Important for real-time use & Does not reflect accuracy & \ding{51} \\
\bottomrule
\end{tabularx}
\caption{Summary of Evaluation Metrics: Types, strengths, limitations, and ground-truth (GT) requirements. Scale-invariant metrics do not depend on the point cloud size. GT-free metrics are suitable for real-world scenarios whereas GT-based metrics are used in simulations.}
\label{tab:appendix_summary_metrics}
\end{table*}

Following the literature and based on the nature of the data, this paper uses several metrics to capture different aspects of registration quality, from geometric accuracy to computational efficiency. Table~\ref{tab:appendix_summary_metrics} presents a summary of these metrics, their types,  strengths, limitations, and ground truth requirements. \\ 

\textbf{Relative Rotation Error (RRE):}
Similar to~\cite{Wang2019PRNet:Registration},~\cite{Wang2019DeepRegistration}, and~\cite{Yew2020RPM-Net:Features}, relative rotational error is defined as the deviation from the ground truth rotation matrices:
\begin{equation}
    \text{RE} = \text{arccos} \big( \frac{Tr(R)-1}{2}\big),
    \label{eq:RE}
\end{equation}
Where \textit{Tr} indicates the trace of a matrix and $R$ is the relative rotation matrix, calculated as 
\begin{equation}
    R = R_{gt}R_{est}^\top,
    \label{eq:R}
\end{equation}
where $R_{gt}$ is the ground truth and $R_{est}$ is the estimated rotation matrices.

This rotation error is not specific to any axes and captures the overall misalignment of the point clouds in the 3D space caused by the difference of the estimated and ground-truth rotation matrices. However, a significant limitation is its inability to account for object symmetries (like a cylinder or a cube). It treats only the labeled ground truth as valid, and incorrectly penalizes equivalent rotations that result in physically identical alignments.

\textbf{Relative Translation Error (RTE):}
Following~\cite{Wang2019PRNet:Registration},~\cite{Wang2019DeepRegistration}, and~\cite{Yew2020RPM-Net:Features}, the relative translation error is considered the Euclidean distance between the ground truth and the estimated translation vectors. 
\begin{equation}
    \text{TE} = \| t_{gt} - t_{est} \|_2.
    \label{er:TE}
\end{equation}
\\
Although rotation and translation errors are commonly reported in research, they do not provide an intuitive understanding of registration quality. Furthermore, they do not support an easy or fair comparison; one approach might get a lower rotation error but a higher translation error than another, or vice versa. As a result, relying only on these two metrics does not provide an objective way to assess the performance of a method~\cite{Fontana2020AAlgorithms, DiLauro2024RobustEvaluation}. In addition, computing these errors requires information about the ground-truth transformation, which is not available in real cases. 

\textbf{Chamfer Distance (CD):}
This metric calculates the average distance between pairs of nearest neighbors of the resulting $\tilde{X}$ and the target $Y$ point cloud~\cite{Bakshi2023ADistance}~\cite{Williams2022PointUtils}:
\begin{align}
\text{CD}(\tilde{X}, Y) =\ & \frac{1}{N} \sum_{i=1}^{N} \|\tilde{x}_i - \text{NN}(\tilde{x}_i, Y)\|_2 \notag \\
& + \frac{1}{M} \sum_{j=1}^{M} \|y_j - \text{NN}(y_j, \tilde{X})\|_2.
\label{eq:chamfer_distance}
\end{align}

Our implementation of the Chamfer distance based on~\cite{Williams2022PointUtils} does not require exact correspondences, as it calculates the distances from each point in a point cloud to its nearest neighbor in the other, rather than its exact corresponding point. Additionally, unlike RRE, it does not unfairly penalize alternative alignments for symmetric objects~\cite{Yew2020RPM-Net:Features}. However, outliers, occlusions, and high point cloud sparsity result in high error values as the distances between points increase. Additionally, this metric is scale-variant, meaning that the same transformation applied to a larger point cloud would result in a different error than if the point cloud were smaller~\cite{Fontana2020AAlgorithms}.

\textbf{Fitness:}
Fitness measures overlapping areas of two point clouds. The better the alignment, the higher the fitness score. This score is defined and implemented as the ratio of the number of inlier correspondences to the total number of points in the target point cloud~\cite{Zhou2018Open3D:Processing}. 
\begin{equation}
    \text{Fitness} = \frac{|\mathcal{I}|}{\text{M}},
    \label{eq:fitness}
\end{equation}
where inlier correspondences ($\mathcal{I}$) refer to pairs of nearest neighbor points whose Euclidean distances are below a predefined threshold $\tau$: 
\begin{equation}
    \mathcal{I} = \{(i,j) | \tilde{x}_i \in \tilde{X}, y_j \in Y, \|R^*\tilde{x}_i + t^* - y_j\|_2 < \tau \},
    \label{eq:inlier_corr}
\end{equation}
where $\tilde{X}$ and $Y$ are the result and target point clouds, respectively. $R^* \in \mathbb{R}^{3\times 3}$ and $t^* \in \mathbb{R}^{3\times 1}$ are the estimated rotation matrix and translation vector applied to the source to create $\tilde{X}$.  

As Eq.\ref{eq:fitness} implies, the maximum achievable fitness score is 1, indicating perfect alignment where every point in the target has a corresponding inlier in the transformed source. This metric is scale-variant because it relies on the number of points. 

\textbf{Inlier RMSE:}
This metric, often reported with Fitness, measures the average alignment error for all inlier correspondences. This is computed as the root mean square of the Euclidean distances between the inlier correspondences~\cite{Zhou2018Open3D:Processing}:
\begin{equation}
    \text{Inlier RMSE} = \sqrt{\frac{1}{|\mathcal{I}|} \sum_{(i, j) \in \mathcal{I}} \left|| R^* \tilde{x}_i + t^* - y_j \right||_2 }.
    \label{eq:inlier_rmse}
\end{equation}

Inlier RMSE should be interpreted in the context of other metrics such as Chamfer Distance and fitness. The reason is that in the case of a failed registration where no inliers are detected ($\mathcal{I} = \emptyset$), the RMSE value becomes zero, which can misleadingly suggest high-quality alignment. 

Similar to Chamfer distance and fitness, the inlier RMSE is scale-dependent. However, these three metrics do not require ground-truth transformations and, therefore, can serve as useful guidelines for the real-world data. Nevertheless, qualitative analysis is necessary to confirm success.

\begin{table*}[t]
\centering
\setlength{\tabcolsep}{7pt}
\begin{tabular}{l | l | ccccccc | c | c}
\toprule
\textbf{Method} & \textbf{Data} & \textbf{Teeth} & \textbf{Lime} & \textbf{Cube} & \textbf{Lego} & \textbf{Eleph.} & \textbf{House} & \textbf{Shoe} & \textbf{SR (\%)} & \textbf{Time} \\
\midrule
\multirow{4}{*}{RPMNet~\cite{Yew2020RPM-Net:Features}} & CD & .205 & .270 & .290 & .218 & .173 & .264 & .191 & \multirow{4}{*}{28.6} & \multirow{4}{*}{0.042s} \\
 & Fit. & 1.00 & 1.00 & 1.00 & 1.00 & 1.00 & 1.00 & .968 & & \\
 & RMSE & .059 & .047 & .023 & .065 & .064 & .132 & .106 & & \\
 & \textbf{Status} & failed & success & success & failed & failed & failed & failed & & \\
\midrule
\multirow{4}{*}{Predator~\cite{Huang2020PREDATOR:Overlap}} & CD & .185 & .270 & .289 & .210 & .158 & .284 & .092 & \multirow{4}{*}{28.6} & \multirow{4}{*}{\textbf{0.038s}} \\
 & Fit. & 1.00 & 1.00 & 1.00 & 1.00 & 1.00 & .992 & 1.00 & & \\
 & RMSE & .048 & .048 & .023 & .055 & .061 & .089 & .059 & & \\
 & \textbf{Status} & failed & success & success & failed & failed & failed & failed & & \\
\midrule
\multirow{4}{*}{GeoTrans.~\cite{Qin2022GeometricRegistration}} & CD & .324 & .260 & .295 & .259 & .184 & .260 & .183 & \multirow{4}{*}{28.6} & \multirow{4}{*}{0.042s} \\
 & Fit. & 1.00 & 1.00 & 1.00 & 1.00 & 1.00 & 1.00 & 1.00 & & \\
 & RMSE & .053 & .041 & .024 & .068 & .067 & .132 & .101 & & \\
 & \textbf{Status} & failed & success & success & failed & failed & failed & failed & & \\
\midrule
\multirow{4}{*}{RegTR~\cite{Yew2022REGTR:Transformers}} & CD & .196 & .270 & .292 & .261 & .190 & .267 & .105 & \multirow{4}{*}{28.6} & \multirow{4}{*}{\textbf{0.038s}} \\
 & Fit. & 1.00 & 1.00 & 1.00 & 1.00 & 1.00 & 1.00 & 1.00 & & \\
 & RMSE & .059 & .047 & .023 & .055 & .068 & .132 & .067 & & \\
 & \textbf{Status} & failed & success & success & failed & failed & failed & failed & & \\
\midrule
\multirow{4}{*}{LoGDesc~\cite{Slimani2024LoGDesc:Registration}} & CD & .186 & .366 & .292 & .207 & .164 & .222 & .092 & \multirow{4}{*}{28.6} & \multirow{4}{*}{0.043s} \\
 & Fit. & 1.00 & .989 & 1.00 & 1.00 & 1.00 & 1.00 & 1.00 & & \\
 & RMSE & .045 & .081 & .024 & .055 & .069 & .052 & .059 & & \\
 & \textbf{Status} & failed & failed & success & failed & failed & success & failed & & \\
\midrule
\multirow{4}{*}{\makecell[l]{\textbf{R3PM-Net (ZS)} \\ \textbf{(ours)}}} & CD & .178 & .326 & .510 & .381 & .169 & .295 & .104 & \multirow{4}{*}{28.6} & \multirow{4}{*}{\underline{0.041s}} \\
 & Fit. & 1.00 & 1.00 & .912 & 1.00 & 1.00 & 1.00 & 1.00 & & \\
 & RMSE & .047 & .060 & .102 & .107 & .070 & .095 & .066 & & \\
 & \textbf{Status} & \textbf{success} & failed & \textbf{success} & failed & failed & failed & failed & & \\
\midrule
\multirow{4}{*}{\makecell[l]{\textbf{R3PM-Net (FT)} \\ \textbf{(ours)}}} & CD & .144 & .288 & .735 & .398 & .167 & .222 & .148 & \multirow{4}{*}{\textbf{42.9}} & \multirow{4}{*}{0.045s} \\
 & Fit. & 1.00 & 1.00 & .795 & 1.00 & 1.00 & 1.00 & 1.00 & & \\
 & RMSE & .034 & .042 & .172 & .124 & .075 & .052 & .096 & & \\
 & \textbf{Status} & \textbf{success} & \textbf{success} & failed & failed & failed & \textbf{success} & failed & & \\
\bottomrule
\end{tabular}
\caption{Complete performance comparison of baseline methods on real-life industrial data. Results are averaged over seven independent runs across seven test objects. A successful registration is defined as achieving accurate alignment, verified visually, in at least four out of seven trials.}
\label{tab:comparative_success_rate}
\footnotesize
\end{table*}

\section{Results} \label{sup:full_results}
The performance of the baseline methods on real-life data is presented in detail in Table~\ref{tab:comparative_success_rate}. The results are averaged across seven independent runs for each of the seven test objects. In this evaluation, a method is considered successful on a case only if it achieves accurate registration in at least four of the seven runs. As mentioned in the paper, the failure or success of the registration is decided based on visual inspection.

\end{document}